%% file: eacl2024_workshop.tex
\pdfoutput=1

\documentclass[11pt]{article}

\usepackage[]{acl}

\usepackage{times}
\usepackage{latexsym}

\usepackage[T1]{fontenc}

\usepackage[utf8]{inputenc}

\usepackage{microtype}

\usepackage{hyperref}
\usepackage{url}
\usepackage{graphicx}
\usepackage{tabularx}
\usepackage{mathtools}
\usepackage{booktabs}
\usepackage{longtable}
\usepackage{tabu}
\usepackage{multirow}
\usepackage{subfigure}
\usepackage{enumitem}
\usepackage{array}
\usepackage{algorithm}
\usepackage{algpseudocode}
\usepackage{arydshln}
\usepackage{wrapfig}
\usepackage{amssymb}

\title{How Does Beam Search improve Span-Level Confidence Estimation in \\Generative Sequence Labeling?}

\author{Kazuma Hashimoto~~~~~Iftekhar Naim~~~~~Karthik Raman \\
Google Research, Mountain View \\
\texttt{\{kazumah, inaim, karthikraman\}@google.com}
}

\begin{document}
\maketitle
\begin{abstract}
Sequence labeling is a core task in text understanding for IE/IR systems.
Text generation models have increasingly become the go-to solution for such tasks (e.g., entity extraction and dialog slot filling).
While most research has focused on the labeling accuracy, a key aspect -- of vital practical importance -- has slipped through the cracks: understanding model confidence.
More specifically, we lack a principled understanding of how to reliably gauge the confidence of a model in its predictions for each labeled span.
This paper aims to provide some empirical insights on estimating model confidence for generative sequence labeling.
Most notably, we find that simply using the decoder's output probabilities \textbf{is not} the best in realizing well-calibrated confidence estimates. 
As verified over six public datasets of different tasks, we show that our proposed approach -- which leverages statistics from top-$k$ predictions by a beam search -- significantly reduces calibration errors of the predictions of a generative sequence labeling model.
\end{abstract}

\input{SEC-intro}

\input{SEC-task}

\input{SEC-method}
\input{SEC-experiments}
\input{SEC-conclusion}

\input{SEC-extra}

\bibliography{custom}
\bibliographystyle{acl_natbib}

\appendix
\input{SEC-appendix}

\end{document}

%% file: SEC-intro.tex
\section{Introduction}

Sequence labeling (e.g., entity extraction) is a fundamental task in building IE/IR systems, such as Web search~\citep{query_seg,web-01,web-02}, QA~\citep{qa-01,qa-02}, and goal-oriented dialog~\citep{converse}.
Prediction confidence is a critical factor for the applications;
it is useful to estimate prediction confidence {\it for each labeled span} in an input text.
Beyond direct application (e.g., knowledge distillation~\citep{hinton2015distilling}), it is crucial to how downstream systems consume the model's output.
For example, a high precision system (say a query parser) may choose to only act on high-confidence spans, while falling back or asking for clarifications for low-confidence ones~\citep{converse}.
Having a well-calibrated model output score -- that correlates well with the correctness of the predictions -- is important for practical adoption.

There have been increasing attempts to apply text generation models to many NLP tasks~\citep{palm2,gpt4}, since the emergence of pre-trained text generation models like GPT~\citep{gpt} and T5~\citep{t5paper}.
Recent work~\citep{athiwaratkun-etal-2020-augmented,seq2seq-slot-filling,karthik-formatting,auto-reg-struct-pre} has shown the advantages of the generative approaches for sequence labeling.
Given the importance of the span-level confidence estimation and the strength of the generative approaches,
\begin{itemize}
    \item[] how do we estimate the confidence of the structured predictions in the text generation?
\end{itemize}
There has been little work on understanding this.

The natural way to estimate a generative model's confidence for each labeled span is via its corresponding token-level posterior probabilities~\citep{asr-conf}.
However, as shown empirically, this approach is not the best;
the posterior probabilities arise solely from the top prediction candidate, which may not capture the underlying uncertainty of the complete decoder distribution.

To overcome this limitation, we propose three methods to take full advantage of top-$k$ statistics given by the beam search; \textbf{AggSpan} aggregates partial span-level probabilities, \textbf{AggSeq} aggregates whole sequence-level probabilities, and \textbf{AdaAggSeq} is an adaptive variant of AggSeq, conditioning on complexity of each input.
Our experiments, comparing the different confidence estimation methods across six diverse datasets and tasks, show that leveraging the beam-search statistics leads to improving model calibration.
Our contributions are summarized as follows:
\begin{itemize}
    \item we propose methods for span-level confidence estimation in generative sequence labeling,
    \item our extensive experiments show the effectiveness of using the beam-search statistics, and
    \item we show the robustness of the AdaAggSeq method with a larger beam size.
\end{itemize}

%% file: SEC-task.tex
\section{Generative Sequence Labeling}

\subsection{Task Description}
\label{subsec:task}

Regardless of what approaches we use, a sequence labeling task $T$ can be formulated as follows:
\begin{equation}
    y = f_T(x),
\end{equation}
where $f_T$ is a task-specific function that takes a text (of $n$ words) $x=[x_1,x_2,\ldots,x_n]$ as an input, and then returns a sequence of $m$ labeled spans $y=[y_1,y_2,\ldots,y_m]$.
We assume that the spans are not nested and not overlapped.
Such a span $y_i$ is a pair of a contiguous word sequence (or a phrase) $s_i$ and its label $\ell_i$: $y_i=(s_i,\ell_i)$.

Here is an example:
\begin{itemize}
    \item[$x$:] [FIFA, World, Cup, 2022, in, Qatar],
    \item[$y$:] [(FIFA, {\tt ASSOCIATION}), (World Cup, {\tt EVENT}), (2022, {\tt YEAR}), (in, {\tt O}), (Qatar, {\tt COUNTRY})],
\end{itemize}
where {\tt ASSOCIATION}, {\tt EVENT}, {\tt YEAR}, and {\tt COUNTRY} are task-specific labels, and {\tt O} is a generic ``{\it outside}'' label that is not any of the task-specific labels.

\subsection{Prediction by Text Generation}

Generative sequence labeling~\citep{NIPS2015-grammer-as,seq2seq-slot-filling} tackles the task by using a conditional text generation model:
\begin{equation}
    \label{eq:argmax}
    y = \underset{y'}{\arg\max}~p_\theta(y'|x),
\end{equation}
where $\theta$ is a set of the model parameters.
The most common approach to the model training is teacher forcing~\citep{teacher-forcing} with human-labeled data, and Equation~(\ref{eq:argmax}) is approximated by using a beam search~\citep{seq2seq}.

In the example in Section~\ref{subsec:task}, $x$ is represented with a list of words and $y$ with a list of position-sensitive phrase-label pairs, but we can use arbitrary text formats as discussed in \citet{karthik-formatting}.
That is, it does not matter which formats we use, as long as we can interpret the outputs.


\subsection{Span-level Confidence Estimation}

The model's predictions are not always correct, and it is practically useful to inspect the model's prediction confidence~\citep{ece1}.
Specifically, this paper focuses on a span-level confidence score:
\begin{equation}
    c_\theta(y_i) \in [0.0, 1.0].
\end{equation}
We can use classifier's output~\citep{bert-calib,bert-ood-robust} with encoder-based token-level classification models~\citep{devlin2018bert}, but it is less trivial in our case.
\citet{auto-reg-uncert} have studied token-level and sequence-level uncertainty estimation in sequence generation tasks; in contrast, we tackle the confidence estimation {\it for each labeled span} consisting of a phrase-label pair and its position.

A straightforward approach is to use the conditional probability as follows:
\begin{equation}
    \label{eq:span_prob}
    c_\theta(y_i) = p_\theta(y_i|x, y_1, \ldots, y_{i-1}),
\end{equation}
which we call ``\textbf{span probability}.''

Assuming that $y_i$ consists of a sequence of $L$ (subword) tokens $[t_{i}^{1},t_{i}^{2},\ldots t_{i}^{L}]$, Equation~(\ref{eq:span_prob}) is computed as follows:
\begin{equation}
    \label{eq:span_prob_detail}
    \prod_{j=1}^{L} p_\theta(t_{i}^{j}|x, y_1, \ldots, y_{i-1}, t_{i}^{1},\ldots, t_{i}^{j-1}).
\end{equation}
Previous work investigated various methods to aggregate partial confidence scores (e.g., token-level scores in ASR systems~\citep{asr-conf} and pixel-level scores in image segmentation~\citep{ece2}), but we have observed that Equation~(\ref{eq:span_prob_detail}) robustly works as a solid baseline.

%% file: SEC-method.tex
\section{Beam Search-based Estimation}

Equation~(\ref{eq:span_prob}) only uses the probability values regarding the top-1 candidate by the beam search.
Therefore, it is not taken into account what labeled spans are likely predicted in other sequence-level outputs in the generative labeling process.

We study two confidence estimation methods that reflect statistical information given by the beam search, inspired by the effectiveness of using the beam search on sequence-level knowledge distillation~\citep{seq-kd-1,seq-kd-2}.

\subsection{Aggregated Span Probability}
\label{subsec:agg_span}

We consider incorporating broader contexts to estimate plausibility of generating $y_i$ given $x$:
\begin{equation}
    \label{eq:agg_span_orig}
    c_\theta(y_i) = p_\theta(y_i|x) = \sum_{z} p_\theta(y_i|x, z) p_\theta(z|x),
\end{equation}
where $z$ is a generated context before predicting $y_i$, and $z=[y_1, \ldots, y_{i-1}]$ is such an example.

Equation~(\ref{eq:agg_span_orig}) is not tractable, and we compute its estimation by the beam search:
\begin{equation}
    \label{eq:agg_span}
    c_\theta(y_i) = \frac{\sum_{z_\mathcal{B}} p_\theta(y_i|x, z_\mathcal{B}) p_\theta(z_\mathcal{B}|x)}{\sum_{z_\mathcal{B}} p_\theta(z_\mathcal{B}|x)},
\end{equation}
where $z_\mathcal{B}$ is a unique context that exists in top-$k$ candidates generated by the beam search.
Note that, if there is only one unique context in the $k$ candidates, Equation~(\ref{eq:agg_span}) is reduced to Equation~(\ref{eq:span_prob}).
We call the method ``\textbf{aggregated span probability}.''

\subsection{Aggregated Sequence Probability}
\label{subsec:agg_seq}

Next, we consider using whole sequence-level information to define $c_\theta(y_i)$, which is a missing ingredient in Equation~(\ref{eq:agg_span}).
More specifically, we aggregate the sequence-level probabilities such that the sequences contain $y_i$:
\begin{equation}
    \label{eq:agg_seq_orig}
    c_\theta(y_i) = \sum_{\hat{y}} p_\theta(\hat{y} | x),
\end{equation}
where $\hat{y}$ is a complete output sequence generated by the model, containing $y_i$.

We use the beam search for its estimation:
\begin{equation}
    \label{eq:agg_seq}
    c_\theta(y_i) = \frac{\sum_{\hat{y}_\mathcal{B}} p_\theta(\hat{y}_\mathcal{B} | x)}{\sum_{j=1}^{k} p_\theta(y^{(j)} | x)},
\end{equation}
where $\hat{y}_\mathcal{B}$ is a $\hat{y}$ that is in the top-$k$ candidates, and $y^{(j)}$ is the $j$-th best candidate.
Intuitively, Equation~(\ref{eq:agg_seq}) counts how frequently $y_i$ appears in the top-$k$ candidates, by weighting the counts with the sequence-level probabilities.
We call the method ``\textbf{aggregated sequence probability}.''
Note that this method is useful only with $k>1$, because $k=1$ always results in $c_\theta(y_i)=1.0$.

\paragraph{Adaptive strategy}
The larger value of $k$ we use, the more output variations this method takes into account, which is expected to be reasonable when the output space is complex.
In contrast, it makes less sense to use a large value of $k$ for an output with only a few non-{\tt O} spans.
To alleviate the potential issue, we propose an adaptive alternative by replacing the constant $k$ in Equation~(\ref{eq:agg_seq}) with an adaptive value $k'\in[2,k]$.
We measure the complexity of the output space by counting the number of non-{\tt O} spans in the top-1 candidate, and set
\begin{equation}
    \label{eq:adaptive_k}
    k'=\max(2, \min(a+b, k)),
\end{equation}
where $a$ is the counted number and $b$ is a hyper-parameter.
We call the method ``\textbf{adaptive aggregated sequence probability}.''

\subsection{Estimation of AggSpan and AggSeq}

In Sections~\ref{subsec:agg_span} and \ref{subsec:agg_seq}, we used the beam search to obtain estimation of Equations~(\ref{eq:agg_span_orig}) and (\ref{eq:agg_seq_orig}), respectively.
We explain the estimation process.

\paragraph{Aggregated span probability:}
We consider the effects of the beam size $k$.
In particular, with $k\gg1$, Equation~(\ref{eq:agg_span}) is expressed as follows:
\begin{eqnarray}
\begin{split}
    &\frac{\sum_{z_\mathcal{B}} p_\theta(y_i|x, z_\mathcal{B}) p_\theta(z_\mathcal{B}|x)}{\sum_{z_\mathcal{B}} p_\theta(z_\mathcal{B}|x)} \\
    \approx &\frac{\sum_{z} p_\theta(y_i|x, z) p_\theta(z|x)}{\sum_{z} p_\theta(z|x)} \\
    = &\sum_{z} p_\theta(y_i|x, z) p_\theta(z|x),
\end{split}
\end{eqnarray}
because of $\sum_{z} p_\theta(z|x)=1$, resulting in Equation~(\ref{eq:agg_span_orig}).

\paragraph{Aggregated sequence probability:}
Similarly, we can express Equation~(\ref{eq:agg_seq}) as follows:
\begin{eqnarray}
\begin{split}
    \frac{\sum_{\hat{y}_\mathcal{B}} p(\hat{y}_\mathcal{B} | x)}{\sum_{j=1}^{k} p(y^{(j)} | x)} \approx &\frac{\sum_{\hat{y}} p_\theta(\hat{y} | x)}{\sum_{y'} p_\theta(y' | x)} \\
    = &\sum_{\hat{y}} p_\theta(\hat{y} | x),
\end{split}
\end{eqnarray}
because of $\sum_{y'} p_\theta(y' | x)=1$, resulting in Equation~(\ref{eq:agg_seq_orig}).

\input{TAB-datasets}

\input{TAB-main-01}

\section{Reliability of Confidence Estimation}
\label{sec:ece}

We expect that, the higher a confidence score is, the more accurate the prediction will be, and vice versa.
To evaluate how reliable the confidence scores are, we adapt a widely-used metric, Expected Calibration Error (ECE)~\citep{ece1,ece2,bert-calib}.

For each evaluation example $x$ in a dataset, we have a prediction $y$ and its corresponding ground-truth annotation $y^*$.
A predicted span in $y$ is treated as correct if it agrees with $y^*$; more concretely, $y^*$ needs to contain a span whose position, phrase, and label are exactly the same as those of the predicted span.
We collect all the predicted spans from all the evaluation examples, resulting in a set of $N$ predicted spans in total.

We then assign a group index $m$ ($1\leq m \leq M$) for each predicted span whose confidence score falls into the $m$-th confidence bin $(\frac{m-1}{M}, \frac{m}{M}]$.
An ECE metric is defined as follows:
\begin{equation}
    \mathrm{ECE} = \frac{1}{N}\sum_{m=1}^{M} N_m |\mathrm{ACC}_m - \mathrm{MC}_m|,
\end{equation}
where $N_m$ is the number of spans in the $m$-th group, and $\mathrm{ACC}_m$ and $\mathrm{MC}_m$ are the accuracy and mean confidence of the group, respectively. We then use the following two ECE metrics:

\noindent
\textbf{- ECE$_\mathrm{ALL}$} evaluates all the predicted spans,

\noindent
\textbf{- ECE$_\mathrm{NO}$} evaluates only non-{\tt O} spans.

%% file: TAB-datasets.tex
\begin{table}[t]

\resizebox{\linewidth}{!}{
\centering
 \begin{tabular}{l|r|r|r||r}
\toprule
    &  Train & Validation & Test & Non-{\tt O} spans \\ \hline
    ATIS & 4,478 & 500 & 893        & 3.4 \\ \hline
    SNIPS & 13,084 & 700 & 700      & 2.6 \\ \hline
    mTOP & 15,667 & 2,235 & 4,386   & 1.7 \\ \hline
    MIT-R & 6,845 & 789  & 1,516    & 2.0 \\ \hline
    NER & 14,987 & 3,466 & 3,684    & 1.8  \\ \hline
    CHUNK & 8,936 & 1,844 & 2,012   & 12.0  \\
    \bottomrule
 \end{tabular}
}
\caption{Statistics of the six datasets.}
   \label{tab:dataset_stats}
\end{table}

%% file: TAB-main-01.tex
\begin{table*}[t]

\resizebox{\textwidth}{!}{
\centering
\begin{tabular}{l|c|c|c|c|c|c}
\toprule
    & \multicolumn{2}{c|}{ATIS~~~(F1: 0.942 $\pm$ 0.003)} & \multicolumn{2}{c|}{SNIPS~~~(F1: 0.930 $\pm$ 0.014)} & \multicolumn{2}{c}{mTOP~~~(F1: 0.906 $\pm$ 0.006)} \\ \hline
                    & ECE$_\mathrm{ALL}$ & ECE$_\mathrm{NO}$ & ECE$_\mathrm{ALL}$ & ECE$_\mathrm{NO}$ & ECE$_\mathrm{ALL}$ & ECE$_\mathrm{NO}$ \\ \hline
 Span    & 0.014 $\pm$ 0.000          & 0.036 $\pm$ 0.001          & \textbf{0.018} $\pm$ 0.003 & 0.039 $\pm$ 0.006                      & 0.026 $\pm$ 0.001 & 0.062 $\pm$ 0.002      \\ \hdashline
 AggSpan & 0.014 $\pm$ 0.000          & 0.036 $\pm$ 0.001          & \textbf{0.018} $\pm$ 0.003 & \textbf{0.038} $\pm$ 0.006 & \textbf{0.025} $\pm$ 0.000 & 0.060 $\pm$ 0.002 \\ \hdashline
 AggSeq  & \textbf{0.011} $\pm$ 0.003 & \textbf{0.020} $\pm$ 0.007 & 0.023 $\pm$ 0.004          & 0.039 $\pm$ 0.007                      & \textbf{0.025} $\pm$ 0.004 & \textbf{0.041} $\pm$ 0.009 \\ \hline \hline

 & \multicolumn{2}{c|}{MIT-R~~~(F1: 0.802 $\pm$ 0.010)} & \multicolumn{2}{c|}{NER~~~(F1: 0.890 $\pm$ 0.010)} & \multicolumn{2}{c}{CHUNK~~~(F1: 0.960 $\pm$ 0.004)} \\ \hline
                    & ECE$_\mathrm{ALL}$ & ECE$_\mathrm{NO}$ & ECE$_\mathrm{ALL}$ & ECE$_\mathrm{NO}$ & ECE$_\mathrm{ALL}$ & ECE$_\mathrm{NO}$ \\ \hline
 Span    & 0.046 $\pm$ 0.003          & 0.119 $\pm$ 0.009          & 0.011 $\pm$ 0.001          & 0.075 $\pm$ 0.004                      & 0.023 $\pm$ 0.001          & 0.026 $\pm$ 0.001          \\ \hdashline
 AggSpan & 0.045 $\pm$ 0.003          & 0.118 $\pm$ 0.009          & 0.010 $\pm$ 0.001          & 0.074 $\pm$ 0.004                      & 0.022 $\pm$ 0.001          & 0.026 $\pm$ 0.001 \\ \hdashline
 AggSeq  & \textbf{0.011} $\pm$ 0.002 & \textbf{0.023} $\pm$ 0.006 & \textbf{0.007} $\pm$ 0.003 & \textbf{0.030} $\pm$ 0.008 & \textbf{0.021} $\pm$ 0.001 & \textbf{0.020} $\pm$ 0.001 \\
\bottomrule
\end{tabular}
}

 \caption{ECE scores ($k=5$) on the ATIS, SNIPS, mTOP, MIT-R, NER, and CHUNK test sets. The lower a score is, the better it is. The value range of the metrics is in $[0.0, 1.0]$. For reference, F1 scores are also shown.}
 \label{tab:test_eval}
\end{table*}

%% file: SEC-experiments.tex
\section{Experiments}

\input{TAB-examples}

\input{TAB-main-02}

We conduct experiments to empirically compare the three methods: \textbf{Span} (Equation~(\ref{eq:span_prob})), \textbf{AggSpan} (Equation~(\ref{eq:agg_span})), and \textbf{AggSeq} (Equation~(\ref{eq:agg_seq})), by setting $k=5$ for the beam search, and $M=10$ for the reliability estimation.
We then evaluate the adaptive AggSeq (\textbf{AdaAggSeq}) with $k=10$.

\subsection{Datasets, Text Format, and Model}
\label{subsec:settings}

To perform the evaluation on diverse datasets and tasks with a strong model, we strictly follow experimental settings in a previous study~\citep{groot}.
The following datasets are used: \textbf{ATIS}~\citep{price1990evaluation}, \textbf{SNIPS}~\citep{coucke2018snips}, \textbf{mTOP}~\citep{li2021mtop}, \textbf{MIT-R},\footnote{\url{https://groups.csail.mit.edu/sls/downloads/}.} \textbf{NER}~\citep{conll-ner}, and \textbf{CHUNK}~\citep{conll-chunk}.
\newline
\noindent
\textbf{- ATIS}: slot-filling in travel assistance,
\newline
\noindent
\textbf{- SNIPS}: slot-filling in virtual assistance,
\newline
\noindent
\textbf{- mTOP}: semantic parsing in voice assistance,
\newline
\noindent
\textbf{- MIT-R}: semantic parsing in dining assistance,
\newline
\noindent
\textbf{- NER}: CoNLL 2003 named entity recognition,
\newline
\noindent
\textbf{- CHUNK}: CoNLL 2000 syntactic chunking.

\noindent
Table~\ref{tab:dataset_stats} shows the number of sentence-level examples, and the average number of (per sentence) annotated non-{\tt O} spans in the validation sets.

We use the ``sentinel+tag (SI)'' format proposed in \citet{karthik-formatting}, to represent the input and output texts in our experiments.
This format is known to be effective in avoiding hallucinations in the text generation.
To run our experiments, we use the pre-trained mT5 ``base''~\citep{mt5paper} in the T5X code base~\citep{roberts2022t5x}.
Details of the fine-tuning process are described in Appendix.

\subsection{Results and Discussions}
\label{subsec:results}

We run all the experiments five times, and the average scores are reported along with the standard deviation values.

\paragraph{Effects of the beam search}

Table~\ref{tab:test_eval} shows the comparison between the three methods.
``Span'' already has a good calibration ability as expected, thanks to the use of the large pre-trained model~\citep{bert-calib}.
We then see that either ``AggSpan'' or ``AggSeq'' is consistently better than ``Span,'' which shows the effectiveness of using the beam search statistics.

\paragraph{Case Study}
We have inspected the results for more intuitive interpretation.
One observation is that ``AggSeq'' tends to better reflect the model's uncertainty when predictions contradict with their ground-truth annotations.
Table~\ref{tab:examples} shows such an example from the MIT-R validation set, where estimated confidence scores are shown for each of the estimation methods.
We can see that ``AggSeq'' assigns the lowest confidence score to the {\tt Location} label, because the ``(in the area, {\tt Location})'' span appears only in the first and fourth candidates out of the top-5 candidates.

\paragraph{Effects of the beam size $k$}
Next, we investigate the effects of the beam size $k$ when using ``AggSeq.''
Table~\ref{tab:test_eval_ada} shows the results with $k=10$, and we can see that $k=10$ performs worse than $k=5$ by comparing the scores with those in Table~\ref{tab:test_eval}.
Only the CHUNK results are comparable; this is presumably because the output space of the CHUNK task is considered to be the most complex as evidenced in Table~\ref{tab:dataset_stats}.

As expected, ``AdaAggSeq'' helps resolve the issue discussed in Section~\ref{subsec:agg_seq}, and the improved scores are even better than those of ``AggSeq'' with $k=5$ (except for CHUNK).
The use of $k'$ in Equation~(\ref{eq:adaptive_k}) makes ``AggSeq'' more robust, and future work is to investigate how to better measure the complexity of each example.

\paragraph{Which one should we use?}
One natural question is which method we want to use in practice; we recommend ``AdaAggSeq'' based on our empirical results, if one can use a validation set to determine the value of $b$.
Otherwise, ``AggSeq'' with $k=5$ is a good choice because it robustly works across the six different datasets.
However, the beam search introduces non-negligible computational costs when performing inference on billions of inputs.
We can use ``Span'' in such a case.

\paragraph{Applicability to blackbox models}
Recently, not all the large pre-trained models are published; GPT-4 ~\citep{gpt4} and PaLM 2~\citep{palm2} are such examples.
In case the token-level probabilities are not visible but the whole sequence-level probabilities are available, (Ada)AggSeq has advantages of being used along with the blackbox models.

%% file: TAB-examples.tex
\begin{table*}[t]

\resizebox{\textwidth}{!}{
\centering
\begin{tabular}{l|l}
\toprule
Input & do you have listings of diners in the area \\ \hdashline
Gold & (do, {\tt O}), (you, {\tt O}), (have, {\tt O}), (listings, {\tt O}), (of, {\tt O}), (diners, {\tt Cuisine}), (in, {\tt O}), (the, {\tt O}), (area, {\tt Location}) \\ \hdashline
\multirow{5}{*}{Top-5} & 1: (do, {\tt O}), (you, {\tt O}), (have, {\tt O}), (listings, {\tt O}), (of, {\tt O}), (diners, {\tt Cuisine}), \underline{(in the area, {\tt Location})} \\
                                   & 2: (do, {\tt O}), (you, {\tt O}), (have, {\tt O}), (listings, {\tt O}), (of, {\tt O}), (diners, {\tt Cuisine}), (in, {\tt O}), (the, {\tt O}), (area, {\tt Location}) \\
                                   & 3: (do, {\tt O}), (you, {\tt O}), (have, {\tt O}), (listings, {\tt O}), (of, {\tt O}), (diners, {\tt Cuisine}), \underline{(in, {\tt Location})}, (the, {\tt O}), (area, {\tt Location}) \\
                                   & 4: (do, {\tt O}), (you, {\tt O}), (have, {\tt O}), (listings, {\tt O}), (of, {\tt O}), \underline{(diners, {\tt O})}, \underline{(in the area, {\tt Location})} \\
                                   & 5: (do, {\tt O}), (you, {\tt O}), (have, {\tt O}), (listings, {\tt O}), (of, {\tt O}), (diners, {\tt Cuisine}), (in, {\tt O}), (the, {\tt O}), \underline{(area, {\tt O})} \\ \hline
Span  & (do, {\tt O})$_\mathbf{0.99}$, (you, {\tt O})$_\mathbf{0.99}$, (have, {\tt O})$_\mathbf{0.99}$, (listings, {\tt O})$_\mathbf{0.99}$, (of, {\tt O})$_\mathbf{0.99}$, (diners, {\tt Cuisine})$_\mathbf{0.99}$, \underline{(in the area, {\tt Location})}$_\mathbf{0.87}$ \\ \hdashline
AggSpan & (do, {\tt O})$_\mathbf{0.99}$, (you, {\tt O})$_\mathbf{0.99}$, (have, {\tt O})$_\mathbf{0.99}$, (listings, {\tt O})$_\mathbf{0.99}$, (of, {\tt O})$_\mathbf{0.99}$, (diners, {\tt Cuisine})$_\mathbf{0.98}$, \underline{(in the area, {\tt Location})}$_\mathbf{0.86}$ \\ \hdashline
AggSeq & (do, {\tt O})$_\mathbf{1.0}$, (you, {\tt O})$_\mathbf{1.0}$, (have, {\tt O})$_\mathbf{1.0}$, (listings, {\tt O})$_\mathbf{1.0}$, (of, {\tt O})$_\mathbf{1.0}$, (diners, {\tt Cuisine})$_\mathbf{0.93}$, \underline{(in the area, {\tt Location})}$_\mathbf{0.63}$ \\
\bottomrule
\end{tabular}
}
 \caption{Confidence estimation for an ambiguous span. Erroneous spans are shown with underlines.}
 \label{tab:examples}
\end{table*}

%% file: TAB-main-02.tex
\begin{table*}[t]

\centering
\small{
\begin{tabular}{l|c|c|c|c|c|c}
\toprule
    & \multicolumn{2}{c|}{ATIS} & \multicolumn{2}{c|}{SNIPS} & \multicolumn{2}{c}{mTOP} \\ \hline
    & ECE$_\mathrm{ALL}$ & ECE$_\mathrm{NO}$ & ECE$_\mathrm{ALL}$ & ECE$_\mathrm{NO}$ & ECE$_\mathrm{ALL}$ & ECE$_\mathrm{NO}$   \\ \hline
 AggSeq    & 0.021 $\pm$ 0.003          & 0.035 $\pm$ 0.009          & 0.036 $\pm$ 0.007          & 0.059 $\pm$ 0.011                & 0.044 $\pm$ 0.005          & 0.068 $\pm$ 0.012      \\ \hdashline
 AdaAggSeq & \textbf{0.009} $\pm$ 0.002 & \textbf{0.016} $\pm$ 0.007 & \textbf{0.016} $\pm$ 0.003 & \textbf{0.028} $\pm$ 0.005 & \textbf{0.013} $\pm$ 0.003 & \textbf{0.022} $\pm$ 0.005 \\ \hline \hline

    & \multicolumn{2}{c|}{MIT-R} & \multicolumn{2}{c|}{NER} & \multicolumn{2}{c}{CHUNK} \\ \hline
    & ECE$_\mathrm{ALL}$ & ECE$_\mathrm{NO}$ & ECE$_\mathrm{ALL}$ & ECE$_\mathrm{NO}$ & ECE$_\mathrm{ALL}$ & ECE$_\mathrm{NO}$   \\ \hline
 AggSeq    & 0.020 $\pm$ 0.002          & 0.028 $\pm$ 0.012          & 0.015 $\pm$ 0.003          & 0.042 $\pm$ 0.011                & 0.023 $\pm$ 0.001          & 0.023 $\pm$ 0.001       \\ \hdashline
 AdaAggSeq & \textbf{0.010} $\pm$ 0.002 & \textbf{0.020} $\pm$ 0.003 & \textbf{0.005} $\pm$ 0.001 & \textbf{0.026} $\pm$ 0.003 & \textbf{0.022} $\pm$ 0.001 & \textbf{0.022} $\pm$ 0.001 \\
\bottomrule
\end{tabular}
}

\caption{Evaluation of ``AggSeq'' and ``AdaAggSeq'' ($k=10$) with $b=3$ for MIT-R and $b=1$ for the rests.}
 \label{tab:test_eval_ada}
\end{table*}

%% file: SEC-conclusion.tex
\section{Conclusion}

We have investigated effective ways of estimating span-level confidence in generative sequence labeling, and shown that the top-$k$ statistics help improve reliability of the estimation.
We believe that our work provides a basis for future work like learning to improve the confidence reliability and using the confidence scores in real applications.

%% file: SEC-extra.tex
\section*{Acknowledgments}
We thank Krishna Srinivasan for his providing useful comments to improve the drat.
We also appreciate fruitful discussions with Leonid Teverovsky and Kiran Yalasangi, about potential use cases of the confidence estimation methods.

\section*{Limitations}

\paragraph{Choice of pre-trained models}
We used a multilingual variant~\citep{mt5paper} of T5~\citep{t5paper} to test the span-level confidence estimation methods, motivated by strong empirical results in previous work~\citep{karthik-formatting,auto-reg-struct-pre}.
However, all the equations in this paper are based on the very basic idea in Equation~(\ref{eq:argmax}), and it is not specific to the T5 model architecture.
For example, the conditional text generation can be implemented with decoder-only models like GPT~\citep{gpt}.
We can use different types of pre-trained text generation models (BART~\citep{bart-paper}, GPT, T5, etc.).

\paragraph{Choice of input/output text formats}
We used a particular input/output text format among a variety of possible formats investigated in previous work~\citep{seq2seq-slot-filling,karthik-formatting}, to minimize concern about hallucinations in the text generation process.
However, all the equations in this paper are not specific to any of the existing text formats.
We can thus adapt the estimation methods to other text formats, as long as we can interpret the outputs as in Section~\ref{subsec:task}.

\paragraph{Application to more complex tasks}
We targeted sequence labeling tasks where labeled spans are not nested as mentioned in Section~\ref{subsec:task}; in other words, there are no overlaps between the labeled spans.
An interesting extension of our work is to adapt the confidence estimation methods to more complex tasks like those in \citet{semeval2022-task10} and \citet{auto-reg-struct-pre}.

\paragraph{Access to prediction probability}
In the end of Section~\ref{subsec:results}, we discussed the applicability of our methods to the blackbox models.
We expect that probability-like scores are available with the predictions, but it would be possible that some APIs only provide predictions without such scores.
Therefore, it is another important line of work to consider reliability of the predictions in such restricted scenarios.

%% file: SEC-appendix.tex
\section*{Appendix}


We fine-tune the pre-trained model for each dataset separately, based on the negative log-likelihood loss~\citep{groot}.
We use the Adafactor optimizer~\citep{adafactor}, along with Z-loss regularization~\citep{z-loss}, where a constant learning rate of 0.001 is used.
The training is run for upto 2500 steps (evaluating checkpoints after every 100 steps).
We select the best checkpoint per the F1 score on the validation set of each dataset.
The T5X code base is publicly available.\footnote{\url{https://github.com/google-research/t5x}.}